\documentclass[a4paper]{article}
\usepackage[margin=1in]{geometry}
\usepackage{amsmath,graphicx,mlspconf}
\usepackage{subfigure}
\usepackage{algorithm}
\usepackage{algorithmic}
\usepackage{xcolor}

\title{CC-Loss: Channel Correlation Loss For Image Classification}
\name{%
    Zeyu~Song,\textsuperscript{1} 
    Dongliang~Chang,\textsuperscript{1}
    Zhanyu~Ma,\textsuperscript{1}
    Xiaoxu~Li,\textsuperscript{2}
    and Zheng-Hua~Tan\textsuperscript{3}
}
\address{%
    \textsuperscript{1} Beijing University of Posts and Telecommunications\\ 
    {\{szy2014, changdongliang, mazhanyu\}@bupt.edu.cn}\\
    \textsuperscript{2} Lanzhou University of Technology\\
    {{xiaoxulilut@gmail.com}}\\
    \textsuperscript{3} Aalborg University\\
    {{zt@es.aau.dk}}
}
\begin{document}
\maketitle

\begin{abstract}
The loss function is a key component in deep learning models. A commonly used loss function for classification is the cross entropy loss, which is a simple yet effective application of information theory for classification problems. Based on this loss, many other loss functions have been proposed,~\emph{e.g.}, by adding intra-class and inter-class constraints to enhance the discriminative ability of the learned features. However, these loss functions fail to consider the connections between the feature distribution and the model structure. Aiming at addressing this problem, we propose a channel correlation loss (CC-Loss) that is able to constrain the specific relations between classes and channels as well as maintain the intra-class and the inter-class separability. CC-Loss uses a channel attention module to generate channel attention of features for each sample in the training stage. Next, an Euclidean distance matrix is calculated to make the channel attention vectors associated with the same class become identical and to increase the difference between different classes. Finally, we obtain a feature embedding with good intra-class compactness and inter-class separability.
Experimental results show that two different backbone models trained with the proposed CC-Loss outperform the state-of-the-art loss functions on three image classification datasets.  
\end{abstract}
\begin{figure*}
    \centering
    \includegraphics[width=0.8\textwidth]{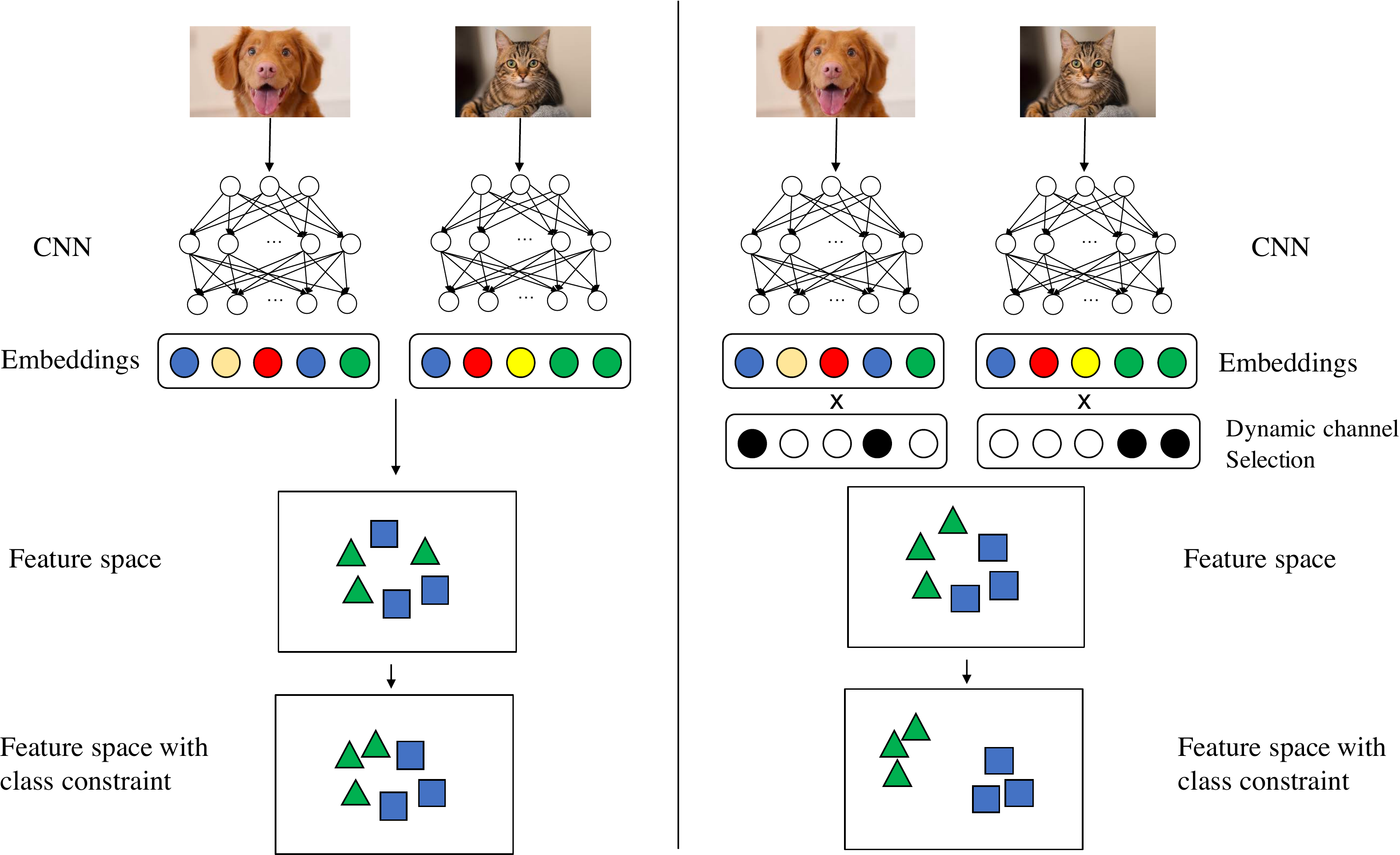}
    \caption{Comparison between the CC-Loss and other related loss functions. The left column shows loss functions that only consider intra- and inter-class constraints. The right column shows the CC-Loss that enhances both the class relation constraints and feature discrimination.}
    \label{fig:insight}
\end{figure*}

\section{Introduction}
\label{sec:intro}

Nowadays, deep learning has become the major methodology in solving computer vision problems. Deep convolutional neural network (CNN)~\cite{krizhevsky2012imagenet,simonyan2014very,huang2017densely,he2016deep} is an important type of deep learning models and has shown outstanding performance on several tasks, including image classification~\cite{deng2009imagenet,du2020fine,zheng2020iu,ma2019fine}, object detection~\cite{lin2014microsoft,kuznetsova2018open}, and object segmentation~\cite{kuznetsova2018open,zhou2017scene}. Image classification, among others, is a basic vision ability of humans. It is a popular deep learning task and has been researched for decades. 

The loss function is a key component in deep CNN models for the purpose of parameter estimation and prior information constraint. For instance, a commonly used loss function, the cross entropy loss (CE-Loss)~\cite{sun2014deep,taigman2014deepface}, measures the distance between two probability distributions which adds information entropy as prior information to classification problem. Training with the CE-Loss improves the classification performance by increasing the predicted Softmax probability of the actual label. It guides the deep learning model to learn separate features for different classes. However, the CE-Loss has two main issues that limit the performance of a CNN model for classification. Firstly, the high level features extracted by CNNs with the CE-Loss are only separable with each other but not discriminative enough~\cite{chen2018virtual}, which can easily lead to over-fitting of the model and thus weak generalization performance. Secondly, the parameters of the deep CNN model are trained jointly with all the classes, which makes the high level features extracted by CNNs to be confused with each other and increases the difficulty of optimization.

Many improvements on loss functions~\cite{chen2018virtual,liu2017sphereface,wen2016discriminative,wang2018cosface,deng2019arcface,hui2019inter,li2019large,chang2020devil} have been made to address the first problem, by introducing constrains to the feature space for discriminative feature embedding. The center loss~\cite{wen2016discriminative} calculates and updates the class centers within each mini batch, which leads to intra-class compactness. The sphereface loss~\cite{liu2017sphereface} introduces an angular margin to optimize the intra-class and inter-class relationship simultaneously, which yields features that are more discriminative. The cosface loss~\cite{wang2018cosface} adds a cosine angular margin according to the target logits to gain better performance against the sphereface loss. The arcface loss~\cite{deng2019arcface} further develops the angular margin to a addictive angular margin, using summation rather than multiplication for efficient training. The inter-class angular loss (ICAL)~\cite{hui2019inter} and the focal inter class angular loss (FICAL)~\cite{wei2019fical} also consider the angular constrains among classes. By using the cosine distances between the categories, these two loss functions both obtained higher classification accuracy among these loss functions. In addition to these improvements on loss functions for image classification, the focal loss~\cite{lin2017focal} is a loss function proposed for object detection, assigning bigger weights to hard examples and smaller weights to easy examples to deal with class imbalance problem.

However, few loss functions have considered the second problem,~\emph{i.e.}, combining the network structure with intra-class and inter-class relationships to decrease the confusion between the features and reduce the difficulty of optimization. To handle this problem and motivated by the mutual channel loss~\cite{chang2020devil}, a recently proposed loss function that fixes the number of deep CNN channels for each class to obtain class-aligned features, we want to make different classes correspond to different channels of the features. 
This can make the samples belonging to different classes to train different parts of the deep CNN model and further learn the features from different classes to be distinguishable from each other. 
To this end, we propose a new loss function to align the channels to each sample. As shown in  Fig.~\ref{fig:insight}, for the samples from the same class, the proposed loss function can make the channel attention vectors be similar to each other. At the meantime, for the samples belong to different classes, the difference between the channel attention vectors will be increased. Finally, we can obtain high intra-class and low inter-class similarity features, which are beneficial for the image classification task.

\begin{figure*}
    \centering
    \includegraphics[width=0.8\textwidth]{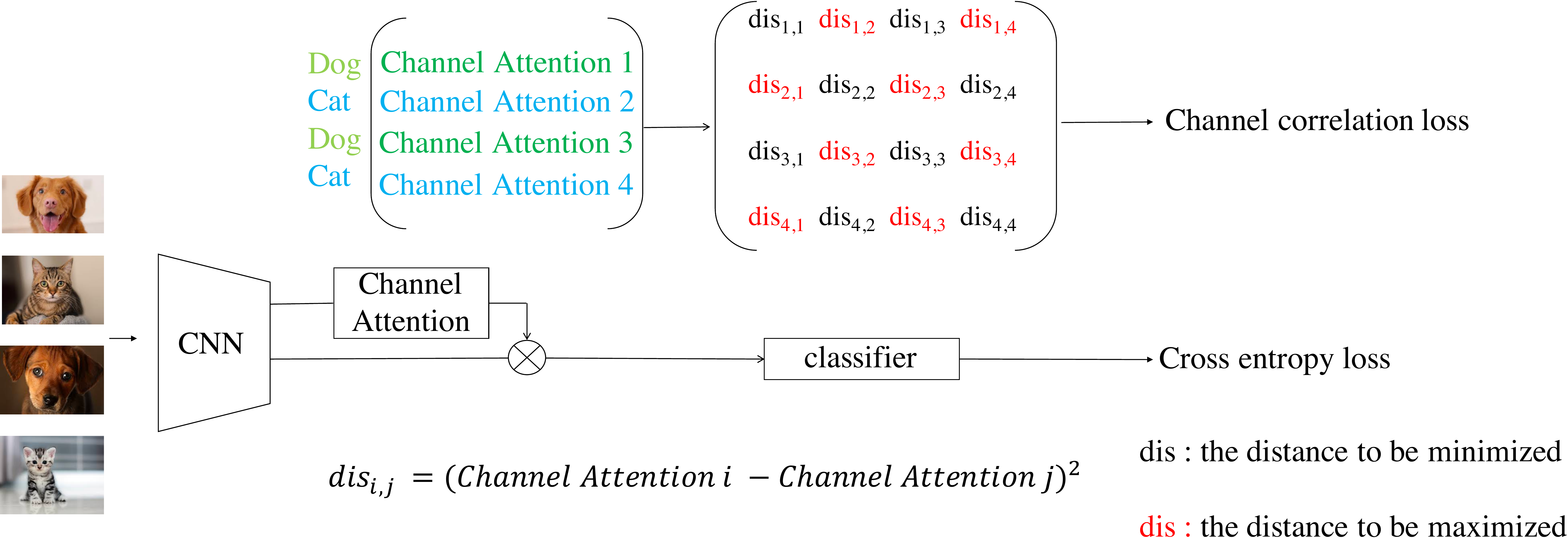}
    \caption{The pipeline of the proposed CC-Loss. Channel attention vectors extracted from each sample are labeled with ground truth classes. The labeled channel attention vectors will then be used to calculate a distance matrix, representing the similarity between channel attentions of different samples. The similarity of samples belonging to the same class should be minimized while the similarity of samples belonging to the different class should be maximized. The distances between samples from the same class is marked in black and those from different classes are marked in red.}
    \label{fig:pipline}
\end{figure*}

The proposed loss function is termed as the channel correlation loss (CC-Loss). As shown in Fig.~\ref{fig:pipline}, it has two components that work together for feature embedding. Firstly, the  squeeze-and-excitation module (SE module)~\cite{hu2018squeeze} is introduced to generate channel attention for each sample, which is a widely used channel attention module in image classification. After that, we need to consider the intra-class and inter-class relations based on the channel attention vectors. Following the previous idea in the loss function design, a loss function needs to minimize the intra-class difference and maximize the inter-class difference.
Therefore, the proposed CC-Loss calculates an Euclidean distance matrix of the attention vectors from one mini-batch.
Each item of the calculated matrix represents the Euclidean distance between the channel attention vectors taken from two input images. By  minimizing the sum of the Euclidean distances from the same class and maximizing the sum of the Euclidean distances from different classes simultaneously, the CC-Loss guided the CNN model to extract the features that have good intra-class compactness and inter-class separability.

The experiments have been conducted by applying the proposed CC-Loss on three commonly used datasets and two network architectures. Experimental results show that the models trained with the CC-Loss is able to extract more discriminative features and outperform the state-of-the-art methods.

\section{The CC-Loss Framework}
\label{sec:meth}
\subsection{Channel attention module (CAM)}
For the purpose of obtaining  the channel attention vectors for each sample, we add the SE module to the classifier. The classifier in this work contains a hidden layer and a classification layer.
The output of the hidden layer is the features, which are the input to the classification layer.
We define the function of the hidden layer as $f_{1}(\cdot)$
and the function of the classification layer as $f_{2}(\cdot)$. Denote the features extracted from the CNN backbone as $\boldsymbol{F} = \left\{F_{1}, F_{2}, \cdots, F_{N} \right\}$, and the output of the classification layer is calculated as $ P_1 = f_{2}(\text{ReLU}(f_{1}(F)))$.

Denote the function of the SE module as $f_{se}(\cdot)$ and the channel attention vector as $\textbf{c} = \left\{c_{1}, c_{2}, \cdots, c_{n}\right\}$ ($n$ is channel numbers of features). The channel attention vector will be multiplied to the output vector of $f_{1}(\cdot)$ element-wise. Then the
calculation process of the classifier will be  $ P_2 = f_{2}(\text{ReLU}(f_{1}(F) \cdot f_{se}(F)))$.



\subsection{Channel Correlation Loss}
The CE-Loss in each mini-batch can be defined as
\begin{equation}
\label{Eq:CEL}
    L_{ce_1} = -\frac{1}{N}\sum_{n=1}^{N}\log(\frac{e^{P_1^{Y_{n}}}}{\sum^{K}_{i=1}e^{P_1^{i}}}),
\end{equation}
where~\textit{N} is the batch size,~\textit{K} is the total number of classes, and $Y_{n}$ is the label of the $n^{th}$ sample. 


In the CC-Loss, we use the channel attention vectors from one mini batch to represent the intra- and inter-class relationships. Hence, the channel attention matrix $\boldsymbol{C}$ is presented as
\begin{equation}
    \boldsymbol{C}=\left[
    \begin{matrix}
        c_{1,1} & c_{1,2} & \cdots & c_{1,D} \\
        c_{2,1} & c_{2,2} & \cdots & c_{2,D} \\
        \vdots & \vdots & \ddots & \vdots \\
        c_{N,1} & c_{N,2} & \cdots & c_{N,D} \\
    \end{matrix}
    \right]_{N \times D},
\end{equation}
where $c_{i,j}$ is the $j^{th}$ element in the channel attention vector of the $i^{th}$ sample. $N$ the is batch size and $D$ is the dimensionality of the hidden layer.

Based on this design, we need to calculate the Euclidean distance between channel attention vectors to obtain the distance matrix from $\boldsymbol{C}$. The distance between the channel attention vector of sample $i$ and sample $j$ is
\begin{equation}
    d_{i,j} = \sum_{d = 1}^{D}(c_{i, d} - c_{j, d})^{2}.
\end{equation}

Now we obtain the distance matrix $\boldsymbol{D}$ of channel attention vectors for one mini-batch as
\begin{equation}
    \boldsymbol{D}=\left[
    \begin{matrix}
        d_{1,1} & d_{1,2} & \cdots & d_{1,N} \\
        d_{2,1} & d_{2,2} & \cdots & d_{2,N} \\
        \vdots & \vdots & \ddots & \vdots \\
        d_{N,1} & d_{N,2} & \cdots & d_{N,N}
    \end{matrix}
    \right]_{N \times N}.
\end{equation}

It is obvious that the Euclidean distance between the same sample will be $d_{i,i} = 0$ and the distance between the same sample pair is symmetric~\emph{i.e.}, $d_{i,j} = d_{j,i}$. Hence, we can further simplify the matrix to an upper triangular matrix as
\begin{equation}
   \boldsymbol{D}^{\text{UT}} = \left[
    \begin{matrix}
        0 & d_{1,2} & \cdots & d_{1,N} \\
        0 & 0 & \cdots & d_{2,N} \\
        \vdots & \vdots & \ddots & \vdots \\
        0 & 0 & \cdots &0
    \end{matrix}
    \right]_{N \times N}.
\end{equation}

Now the distance matrix is simple enough to calculate the intra-class component and the inter-class component for the CC-Loss. The intra-class component is the sum of the distances within the same class. For example, if sample $i$ and sample $j$ are from the same class, $d_{i,j}$ is added to the intra-class component. Hence, $L_{intra}$ is calculated as
\begin{equation}
\label{Eq:Intra Comp}
    L_{intra} = \sum_{i=1}^{N}\sum_{j=i}^{N}d_{i,j}\cdot \text{I}(Y_{i}, Y_{j}),
\end{equation}
where the operator $\text{I}(Y_{i},Y_{j})$ is defined as \begin{equation}
\nonumber
\text{I}(Y_{i},Y_{j})=\left\{
\begin{split}
    1, &\ \ \ Y_{i}=Y_{j}\\
    0, &\ \ \  \text{otherwise}
\end{split}
\right. .
\end{equation} 
The inter-class component is the sum of distances from different classes. Therefore, $L_{inter}$ is
\begin{equation}
\label{Eq:Inter Comp}
    L_{inter} = \sum_{i=1}^{N}\sum_{j=i}^{N}d_{i,j}\cdot \left[1-\text{I}(Y_{i},Y_{j})\right].
\end{equation}
The distances between the same class should be minimized, which forces the channel selection to be the same and enhances the intra-class compactness. Meanwhile, the distances among different classes should be maximized, which forces the channel selection to be different and enlarges the inter-class separability. Hence, the final loss function is defined as
\begin{equation}
    L = L_{ce_2} + \lambda\cdot \frac{L_{intra}}{L_{inter} + \epsilon},
\end{equation}

\begin{equation}
    L_{ce_2} = -\frac{1}{N}\sum_{n=1}^{N}\log(\frac{e^{P_2^{Y_{n}}}}{\sum^{K}_{i=1}e^{P_2^{i}}}),
\end{equation}
where $\epsilon$ is a small positive constant and $\lambda$ is a hyper-parameter.

\subsection{Computational Complexity Reduction}
The computational complexity to compute the distance matrix $D$ is $O(N^{2})$, while it is $O(N^{2})$ for calculating $L_{intra}$ and $L_{inter}$. Therefore, the overall computational complexity for CC-Loss is $O(N^{2})$. In order to facilitate the calculation, we reduce the computational complexity of calculating the distance matrix from $O(N^{2})$ to $O(N)$ by the following steps. 

Each item $d_{i,j}$ in the distance matrix $D$ is calculated as,
\begin{equation}
\begin{split}
    d_{i,j} = &\sum_{k = 1}^{D}(c_{i,k} - c_{j,k})^{2}\\
  = &\sum_{k = 1}^{D}c_{i,k}^{2} + \sum_{k = 1}^{D}c_{j,k}^{2} - 2\sum_{k = 1}^{D}c_{i,k}c_{j,k}.
  \end{split}
\end{equation}
Donate $s_{i}$ as the square sum of each item of the $i^{th}$ channel attention vector in $C$, which is given as, 
\begin{equation}
    s_{i} = \sum_{k = 1}^{D}c_{i,k}^{2},
\end{equation}
we and then calculate two auxiliary matrix $E$ and $E^\text{T}$ with computational complexity $O(n)$. Here, matrix $E$ is defined as,
\begin{equation}
    E=\left[
    \begin{matrix}
        s_{1} & s_{1} & \cdots & s_{1} \\
        s_{2} & s_{2} & \cdots & s_{2} \\
        \vdots & \vdots & \ddots & \vdots \\
        s_{N} & s_{N} & \cdots & s_{N}
    \end{matrix}
    \right]_{N \times N}.
\end{equation}
After that, we calculate an auxiliary matrix $G = C \times C^\text{T}$ with $O(n)$, which is
\begin{equation}
\resizebox{0.85\linewidth}{!}{$
  G=  \left[
    \begin{matrix}
        c_{1,1} & c_{1,2} & \cdots & c_{1,D} \\
        c_{2,1} & c_{2,2} & \cdots & c_{2,D} \\
        \vdots & \vdots & \ddots & \vdots \\
        c_{N,1} & c_{N,2} & \cdots & c_{N,D} \\
    \end{matrix}
    \right] \times
    \left[
    \begin{matrix}
        c_{1,1} & c_{2,1} & \cdots & c_{N,1} \\
        c_{1,2} & c_{2,2} & \cdots & c_{N,2} \\
        \vdots & \vdots & \ddots & \vdots \\
        c_{1,D} & c_{2,D} & \cdots & c_{N,D} \\
    \end{matrix}
    \right]
    $}.
\end{equation}
Note that $G$ is a $N \times N$ matrix and the item $g_{i,j}$ in $G$ is
\begin{equation}
    g_{i,j} = \sum_{k = 1}^{D}c_{i,k}c_{j,k}.
\end{equation}

Then, the matrix $D$ is now calculated as $D = E + E^\text{T} - 2G$. The overall computational complexity is reduced to $O(N)$.


\section{Experimental Results and Discussions}
\label{sec:exp}
In this section, the proposed CC-Loss is evaluated on three datasets: MNIST~\cite{deng2012mnist}, CIFAR-$100$~\cite{krizhevsky2009cifar}, and Cars-$196$~\cite{KrauseStarkDengFei-Fei_3DRR2013}. Two CNN backbones, VGG$16$ and ResNet$18$, are investigated. Furthermore, we compare the CC-Loss with other loss functions including the CE-Loss~\cite{sun2014deep}, the Focal loss~\cite{lin2017focal}, the A-softmax loss~\cite{liu2017sphereface}, the inter-class angular loss (ICAL)~\cite{hui2019inter}, and the focal inter class angular loss (FICAL)~\cite{wei2019fical}. 

\subsection{Datasets description}
\textbf{MNIST}~\cite{deng2012mnist} is a handwritten digits classification dataset contains numbers from zero to nine. It includes a training set with $60,000$ images and a test set with $10,000$ images. All the images have been normalized and resized to the size of $28\times28$.\\
\textbf{CIFAR-$100$}~\cite{krizhevsky2009cifar} is a natural scene image classification dataset that contains $60,000$ colored images from $100$ classes. The training set and the test set contain $50,000$ and $10,000$ images, respectively. All the images are with the resolution of $32\times32$. \\
\textbf{Cars-196}~\cite{KrauseStarkDengFei-Fei_3DRR2013} is a fine-grained vehicle classification dataset, which contains $8,144$ training images and $8,041$ test images from $196$ classes. 
\subsection{Implementation details}
We applied the CC-loss function to two widely used CNN architectures,~\emph{i.e.}, VGG$16$ and ResNet$18$.

\begin{table*}
    \centering
    \caption{Classification accuracy($\%$) with different loss functions. The backbone models were VGG$16$ and ResNet$18$ and evaluated on three datasets. All the CC-loss results are the average of five rounds evaluations.}
    \resizebox{0.9\textwidth}{!}{
    \begin{tabular}{ccccc}
    \hline
    Loss Function & Backbone Model & MNIST & CIFAR-$100$ & Cars-$196$ \\
    \hline
    CE Loss & VGG$16$/ResNet$18$ & $97.43$/$97.52$ & $74.49$/$77.38$ & $88.02$/$85.77$ \\
    Focal Loss & VGG$16$/ResNet$18$ & $97.64$/$97.68$ & $74.46$/$77.63$ & $88.21$/$85.98$ \\
    A-softmax Loss & VGG$16$/ResNet$18$ & $98.10$/$98.32$ & $74.55$/$77.78$ & $90.02$/$87.22$ \\
    MC Loss & VGG$16$/ResNet$18$ & $98.20$/$\underline{98.45}$ & $72.51$/$70.18$ & $\textbf{92.80}$/- \\
    ICAL & VGG$16$/ResNet$18$ & $97.83$/$98.21$ & $74.79$/$77.71$ & $89.32$/$86.67$ \\
    FICAL & VGG$16$/ResNet$18$ & $\underline{98.22}$/$98.40$ & $\underline{74.98}$/$\underline{78.18}$ & $89.70$/$\underline{87.38}$ \\
    \hline
    CC-Loss & VGG$16$ + CAM/ResNet$18$ + CAM & $\textbf{98.32}$ $\pm$ $\textbf{0.08}$/$\textbf{98.52}$ $\pm$ $\textbf{0.09}$ &  $\textbf{75.49}$ $\pm$ $\textbf{0.15}$/$\textbf{78.23}$ $\pm$ $\textbf{0.07}$ & $\underline{91.46}$ $\pm$ $\underline{0.09}$/$\textbf{88.41 $\pm$ 0.06}$ \\
    \hline
    \end{tabular}}
    \label{tab:SOTA}
\end{table*}


 All the methods were trained with the stochastic gradient descent (SGD)~\cite{bottou2010large} algrithm with $300$ epochs. The batch sizes for MNIST, CIFAR-$100$, and Cars-$196$ were set to $32$, $32$,  and $16$, respectively. The input image sizes were $24\times24$, randomly cropped $32\times32$, and randomly cropped $224\times224$ following a zero padding with size $4$ for MNIST, CIFAR-$100$, and Cars-$196$, respectively. Horizontally flipping with $0.5$ probability was also applied when training CIFAR-$100$ and Cars-$196$. The initial learning rate was set to $0.1$ and adjusted by the cosine annealing schedule~\cite{huang2017snapshot} to $1e-5$. We set the weight decay as $5e-4$ and the momentum as $0.9$. When training MNIST and CIFAR-$100$, the network parameters were initialized with the method in~\cite{he2015delving}. While training Cars-$196$, we used the ImageNet pretrained parameters as the initial ones. For the hyper-parameters of the loss functions used for comparison, we followed the referred papers. For the CC-Loss, we set $\lambda$ to $1$. All the experiments results reported shared the same hyper-parameters.
 
 We conducted each of our experiments five rounds and report the average accuracy and margin in the form of $accuracy \pm margin$. The margin is defined as maximum absolute difference between the five results and it's average. For baseline methods in Table~\ref{tab:SOTA} and Table~\ref{tab:CAM}, we refer to results from~\cite{chang2020devil} and~\cite{wei2019fical}, which only report average values.
\begin{table}
    \centering
    \caption{Ablation Study on the batch size for the CC-Loss. The experiments were conducted on CIFAR-$100$ as well as Cars-$196$ datasets with VGG$16$ backbone.}
    \vspace{2mm}
    \begin{tabular}{ccc}
    \hline
    Batch Size & CIFAR-$100$ & Cars-$196$\\
    \hline
    $8$ & $74.9 \pm 0.12$ & $91.3 \pm 0.12$ \\
    $16$ & $75.3 \pm 0.07$ & $91.5 \pm 0.09$ \\
    $32$ & $75.5 \pm 0.15$ & $91.3 \pm 0.16$ \\
    $64$ & $75.1 \pm 0.14$ & $90.6 \pm 0.18$ \\
    $128$ & $74.5 \pm 0.14$ & $89.9 \pm 0.15$ \\
    \hline
    \end{tabular}
    \label{tab:batchsize}
\end{table}

\begin{table}
    \centering
    \caption{The $p$ values of one sample $t$-test.}
    \resizebox{0.45\textwidth}{!}{
    \begin{tabular}[width=0.5\textwidth]{cccc}
    \hline
    Base Model & MNIST & CIFAR-$100$ & Cars-$196$ \\
    \hline
    VGG$16$ & $3.9\times10^{-3}$ & $4.7\times10^{-4}$ & $2.1\times10^{-7}$ \\
    ResNet$18$ & $3.3\times10^{-3}$ & $3.8\times10^{-3}$ & $4.3\times10^{-7}$ \\
    \hline
    \end{tabular}}
    \label{tab:t-rest}
\end{table}

\begin{table*}
    \centering
    \caption{Ablation study of the channel attention module.}
    \resizebox{0.9\textwidth}{!}{
    \begin{tabular}{ccccc}
    \hline
    Loss Function & Backbone Model & MNIST & CIFAR-$100$ & Cars-$196$ \\
    \hline  
    CE Loss & VGG$16$/ResNet$18$ & $97.43$/$97.52$ & $74.49$/$77.38$ & $88.02$/$85.77$ \\
    CE Loss & VGG$16$ + CAM/ResNet$18$ + CAM & $97.66$ $\pm$ $0.02$/$97.67$ $\pm$ $0.01$ & $74.54$ $\pm$ $0.02$/$77.57$ $\pm$ $0.01$ & $88.15$ $\pm$ $0.03$/$86.04$ $\pm$ $0.02$ \\
    CC Loss & VGG$16$ + CAM/ResNet$18$ + CAM & $98.32$ $\pm$ $0.08$/$98.52$ $\pm$ $0.09$ &  $75.49$ $\pm$ $0.15$/$78.23$ $\pm$ $0.07$ & $91.46$ $\pm$ $0.09$/$88.41$ $\pm$ $0.06$ \\
    \hline
    \end{tabular}}
    \label{tab:CAM}
\end{table*}

\begin{table}
    \centering
    \caption{The effect of $\lambda$ in CC-Loss. The experiments were conducted using VGG$16$ + CAM as the backbone }
    \vspace{2mm}
    \resizebox{0.45\textwidth}{!}{
    \begin{tabular}[width=0.5\textwidth]{cccc}
    \hline
    $\lambda$ & MNIST & CIFAR-$100$ & Cars-$196$ \\
    \hline
    $0$ & $97.66$ $\pm$ $0.02$ & $74.54$ $\pm$ $0.02$ & $88.15$ $\pm$ $0.03$ \\
    $0.5$ & $98.04$ $\pm$ $0.05$ & $75.12$ $\pm$ $0.03$ & $89.53$ $\pm$ $0.07$ \\
    $1.0$ & $98.32$ $\pm$ $0.08$ & $75.49$ $\pm$ $0.15$ & $91.46$ $\pm$ $0.09$ \\
    $1.5$ & $97.95$ $\pm$ $0.06$ & $75.04$ $\pm$ $0.09$ & $89.42$ $\pm$ $0.05$ \\
    $2.0$ & $97.53$ $\pm$ $0.03$ & $74.32$ $\pm$ $0.12$ & $87.84$ $\pm$ $0.02$ \\
    \hline
    \end{tabular}}
    \label{tab:lambda}
\end{table}

\begin{figure*}
    \centering
    \subfigure[MNIST visualization for CE-Loss]{
    \includegraphics[width=0.45\textwidth]{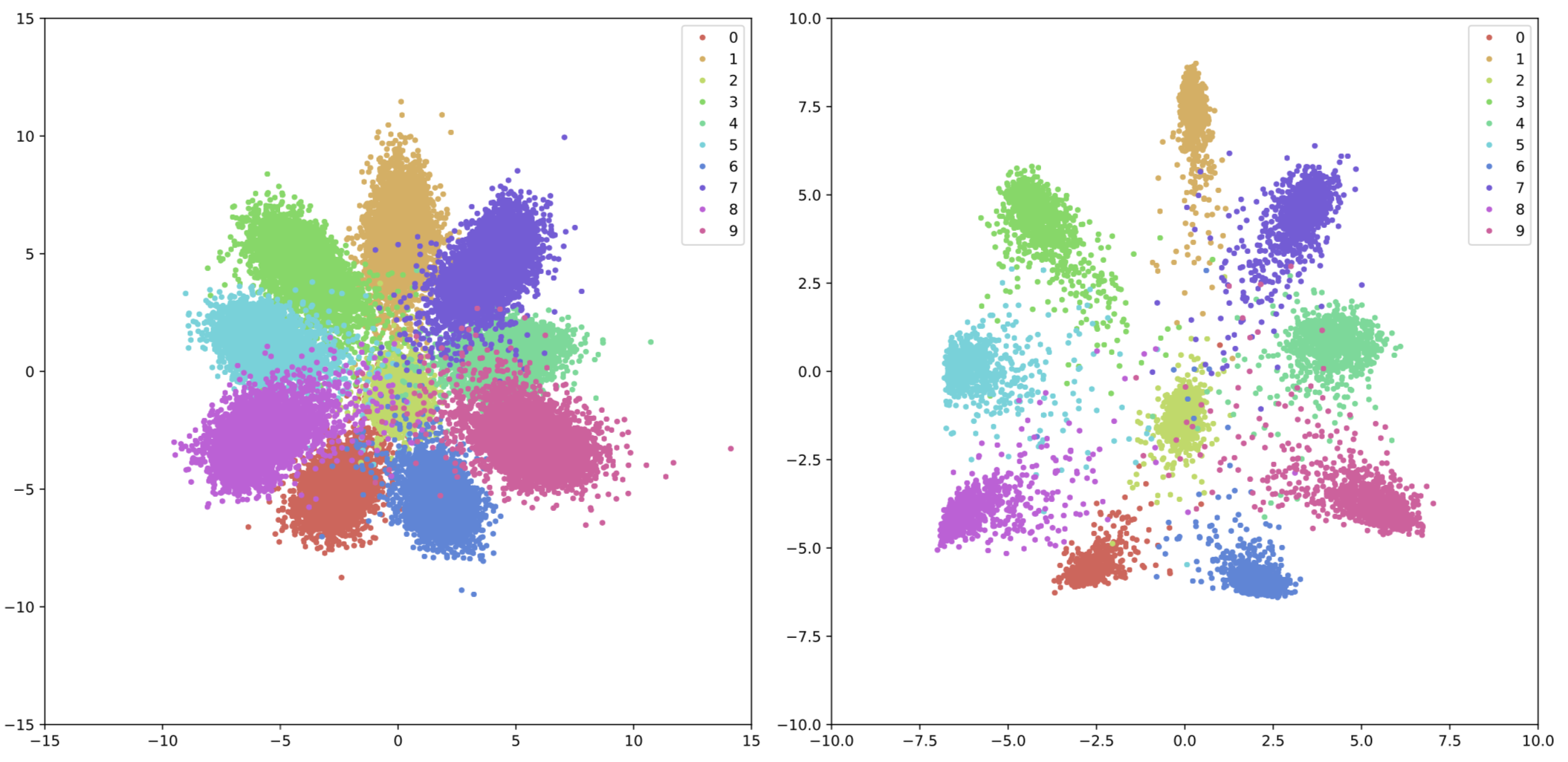}
    }
    \quad
    \subfigure[MNIST visualization for CC-Loss]{
    \includegraphics[width=0.45\textwidth]{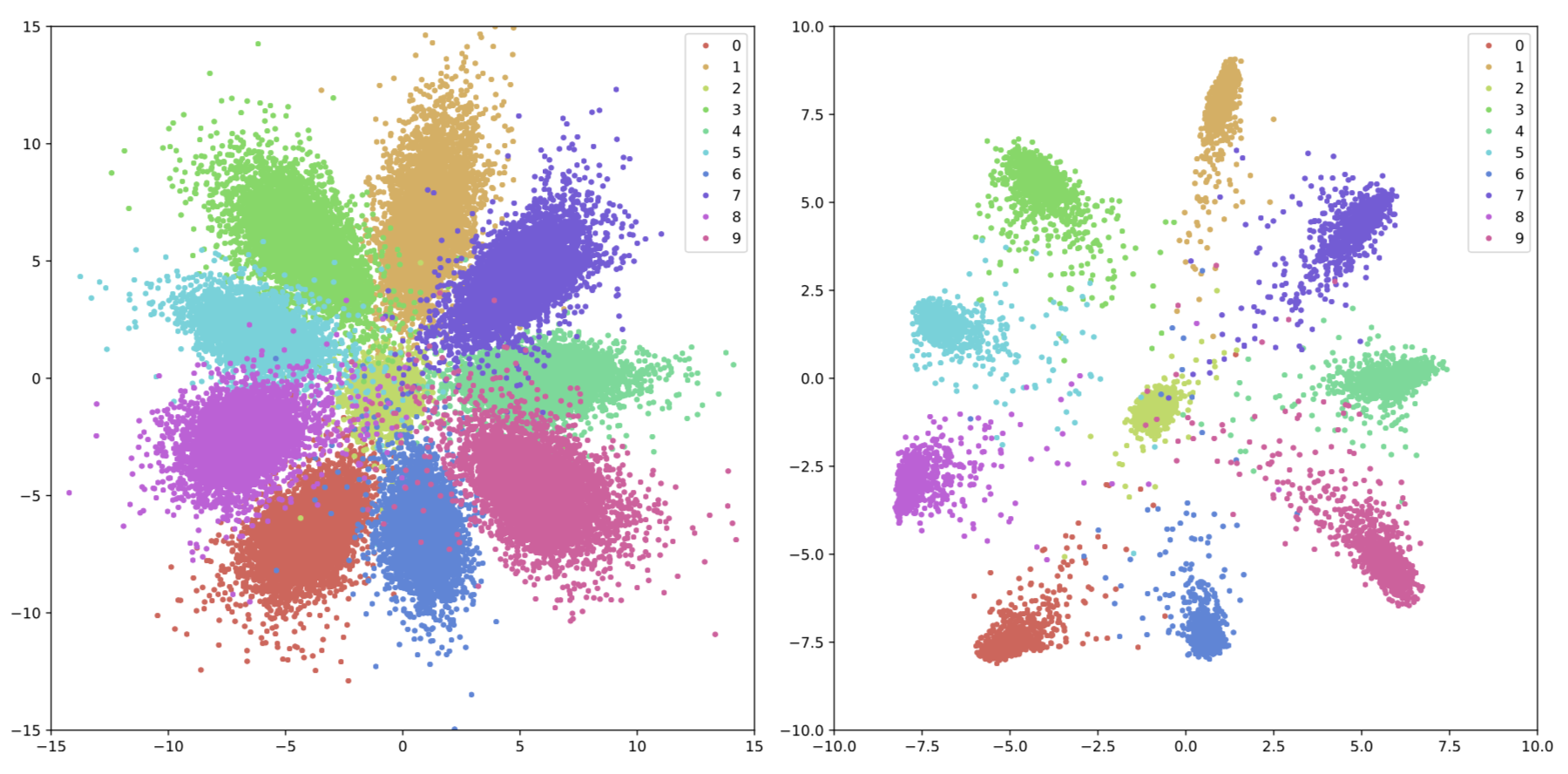}
    }
    \caption{Visualization on MNIST dataset trained by CE-Loss and CC-Loss. The left sub-figure and the right sub-figure shows the results on training set and test set respectively.}
    \label{fig:vis}
    \vspace{-2mm}
\end{figure*}

\vspace{-4mm}
\subsection{Comparison with state-of-the-art methods}
As shown in Table~\ref{tab:SOTA}, the proposed CC-Loss outperforms the CE-Loss by a large margin. For ResNet$18$, the CC-Loss improves the performance from $97.52\%$ to $98.52\%$, $77.38\%$ to $78.23\%$, and $85.77\%$ to $88.41\%$ on MNIST, CIFAR-$100$, and Cars-$196$ dataset, respectively. For VGG$16$, the CC-Loss outperforms the CE-Loss by $0.89\%$, $1\%$, and $3.33\%$ on MNIST, CIFAR-$100$, and Cars-$196$, respectively. These results indicate the class relation constraints in the CC-Loss indeed help the CNN model with more discriminative features, which shows the effectiveness of the CC-Loss. The improvement on Cars-$196$ dataset is the most significant as the class relationship in the fine-grined image classification dataset is more complex. 
Therefore, it benefits more from the class relation constrains. The VGG$16$ backbone gets more improvement over ResNet$18$ for higher parameter numbers bring to higher model capacity, which react better with more prior information generated by CC-Loss. While comparing with the state-of-the-art loss functions that only focus on intra- and inter-class relations, the CC-Loss also shows competitive performance on all the three datasets. For ResNet$18$, CC-Loss gets slightly better results on MNIST and CIFAR-$100$ compared to the second best FICAL loss, and get $1.03\%$ boosting performance on the Cars-$196$ dataset. For VGG-$16$, the CC-Loss gets $0.1\%$ and $0.51\%$ improvement against the FICAL loss on MNIST and CIFAR-$100$. It also gets $1.44\%$ accuracy improvement compared with the second best A-softmax Loss. These demonstrate that the additional dynamic class-channel mechanism can help with the classification task.

Specifically, we further analyze the results of the MC-Loss, another loss function which focuses on channel-wise intra-class and inter-class relationship designed for the fine-grained visual classification (FGVC) task. The MC-Loss has better performance on the Cars-196 dataset, which is a FGVC dataset. But it has worse performance on image classification datasets,~\emph{i.e.}, MNIST and CIFAR-100. The key reason is that MC-Loss allocates constant channel numbers for each class (3 channels for each class in Cars-196) and learns three different feature maps for each class. While in general image classification tasks, there might not be enough separate feature map for each sample, this will yield sub-optimization of the feature maps and can not meet the separation constraint. Furthermore, using constant channel number is unsuitable for CIFAR and MNIST classes, which also causes a worse result of MC-Loss. 

\subsection{The effect of CAM}
We add a channel attention model on top of the CNN backbone to meet the requirement of dynamic channel selection. An ablation study is proposed to show the performance of the channel attention module when CC-Loss is not applied. As shown in Table~\ref{tab:CAM}, in the case of w/o CC-Loss, dynamic channel selection gives slight improvement against the original backbone. When CC-Loss is used, it provides extra class relation constraint to boost the performance. Also, it is worth to note that CC-Loss can not be implemented without the CAM module. 

\subsection{The effect of weight in CC-Loss }
As shown in Table~\ref{tab:lambda}, we evaluated the performance of different $\lambda$, representing the influence of CC-Loss. The best result is obtained when $\lambda$ is $1$. Further increasing $\lambda$ will cause performance drop because we need class-label relationship provided by the CE-Loss.

\subsection{Hyper-parameter tuning on batch size}
Since we optimize the channel attention distances within mini-batchs, the batch size is an important hyper-parameter to be tuned. We analyzed the affect of batch size on CIFAR-$100$ and Cars-$196$ datasets with VGG$16$ as the backbone. As shown in Table~\ref{tab:batchsize}, a smaller batch size leads to a higher classification accuracy, since the class relationship in the distance matrix is simpler and easier to optimize. However, when the batch size tends to be very small, the class relationship will disappear and thus the accuracy is decreased. Therefore, we empirically chose a batch size of $32$ for the CIFAR-$100$ dataset and a batch size of $16$ for the Cars-$196$ dataset.

\vspace{-2mm}
\subsection{Statistical significance analysis}
As the number of experiments is less than $30$, we used the one sample unpaired  $t$-test~\cite{posten1979robustness} to evaluate the statistical significance of the results. We took the values from the state-of-the-art method as the general average value for test. More specifically, we used the results from the A-softmax Loss for Cars-$196$ dataset on VGG$16$ backbone and the results from the FICAL Loss for the rest.
As shown in Table~\ref{tab:t-rest}, all the $p$-values of the one sample $t$-test with SOTA results are below $5\times10^{-3}$, which indicates statistically significant improvements obtained by the CC-Loss compared with the baseline loss functions. 

\vspace{-2mm}
\subsection{Visualization}
We carry out the visualization experiments by training VGG$16$ on the MNIST dataset. The output dimensionality of the first linear layer in VGG$16$ was set to $2$. The features extracted from the first linear layer are visualized to present the feature space obtained by  the CE-Loss and the CC-Loss. From Fig.~\ref{fig:vis}, it can be observed that the features from the CC-Loss are more compact within the same class while more separable among the neighbour classes. This demonstrates the effectiveness of intra-class and inter-class component of CC-Loss.

\vspace{-4mm}
\section{Conclusions}
In this paper, we proposed a new loss function, namely the Channel Correlation Loss (CC-Loss), for the task of image classification. The CC-Loss dynamically selects the CNN channels for each class via the channel attention mechanism. It assigns larger numbers of channels to the harder classes and smaller numbers of channels to the easier class. Furthermore, by considering the Euclidean distances of the channel attention vectors in mini-batches, the CC-Loss is able to maximize both the intra-class compactness and the inter-class separability, which can extract more discriminative features. Experimental results on three datasets with two different backbones demonstrated that the proposed method outperforms commonly used loss functions.


\bibliographystyle{IEEEbib}
\bibliography{main.bib}

\end{document}